# Soil Moisture Estimation in Precision Agriculture: A Knowledge Distillation Approach

**Mohammed Rakib**
Oklahoma State University
mohammed.rakib@osktate.edu

**Vrinda Nagula**
Aurora High School
vnagula07@gmail.com

**Arunkumar Bagavathi**
Oklahoma State University
abagava@okstate.edu

**ABSTRACT (REQUIRED)**

Accurate soil moisture estimation is critical for precision agriculture, influencing irrigation, fertilization, and harvesting strategies. Traditional methods primarily rely on meteorological and soil property data, while recent approaches incorporate visual modalities such as aerial and geospatial imagery. However, these high-resolution images are expensive and computationally demanding to analyze. In this work, we propose Soil Moisture Distillation (SMD), a Knowledge Distillation (KD)-based multimodal learning framework that leverages unimodal teachers to enhance multimodal soil moisture estimation. SMD surpasses not only traditional multimodal fusion techniques, including MIS-ME, but also outperforms unimodal models, demonstrating the benefits of guided unimodal knowledge transfer. Additionally, we conduct a comparative analysis of soil patches and crop images, showing that crop images provide superior visual cues, leading to improved prediction accuracy. SMD achieves a 2% improvement in MAPE for soil patches and 1.2% for crops over state-of-the-art baselines, highlighting the effectiveness of leveraging both meteorological and visual data through knowledge distillation for enhanced soil moisture forecasting.

## INTRODUCTION

Soil moisture plays a vital role in precision agriculture, influencing irrigation strategies, crop health monitoring, and overall farm productivity. Accurate soil moisture estimation helps farmers optimize water usage, reducing waste while ensuring crops receive adequate hydration. Traditionally, soil moisture is measured using in-situ sensors, which provide precise readings but are expensive to deploy and maintain at scale (Phillips *et al.*, 2014; Feng *et al.*, 2023). As an alternative, machine learning-based predictive models leverage meteorological and soil property data to estimate soil moisture, offering a scalable and cost-effective solution (Orth and others, 2021; Filipović *et al.*, 2022). However, relying solely on numerical meteorological data may not be sufficient, as soil moisture is also influenced by physical soil conditions and crop health, which are better captured through visual cues.

Recent research has explored image-based soil moisture estimation, utilizing satellite, aerial, and multispectral imagery to enhance predictive performance (Bertalan *et al.*, 2022; Singh and Gaurav, 2023). While these high-resolution images capture large-scale environmental features, they are expensive to acquire, computationally intensive to process, and often lack fine-grained details necessary for localized soil moisture estimation. To address this, researchers have investigated using ground-level images of soil patches and crops, which offer micro-level insights into soil texture, vegetation status, and water retention (Rakib *et al.*, 2024). However, the effectiveness of such visual information depends on how well it complements traditional meteorological data in a multimodal setting.

Despite the promise of multimodal approaches, integrating heterogeneous data sources presents significant challenges (Du *et al.*, 2023; Zhang *et al.*, 2024). One key issue is modality imbalance (Peng *et al.*, 2022), where dominant modalities can overshadow weaker modalities, leading to suboptimal feature fusion. Traditional multimodal learning models struggle to effectively balance and utilize information from different sources, often prioritizing the most predictive modality while neglecting complementary signals.

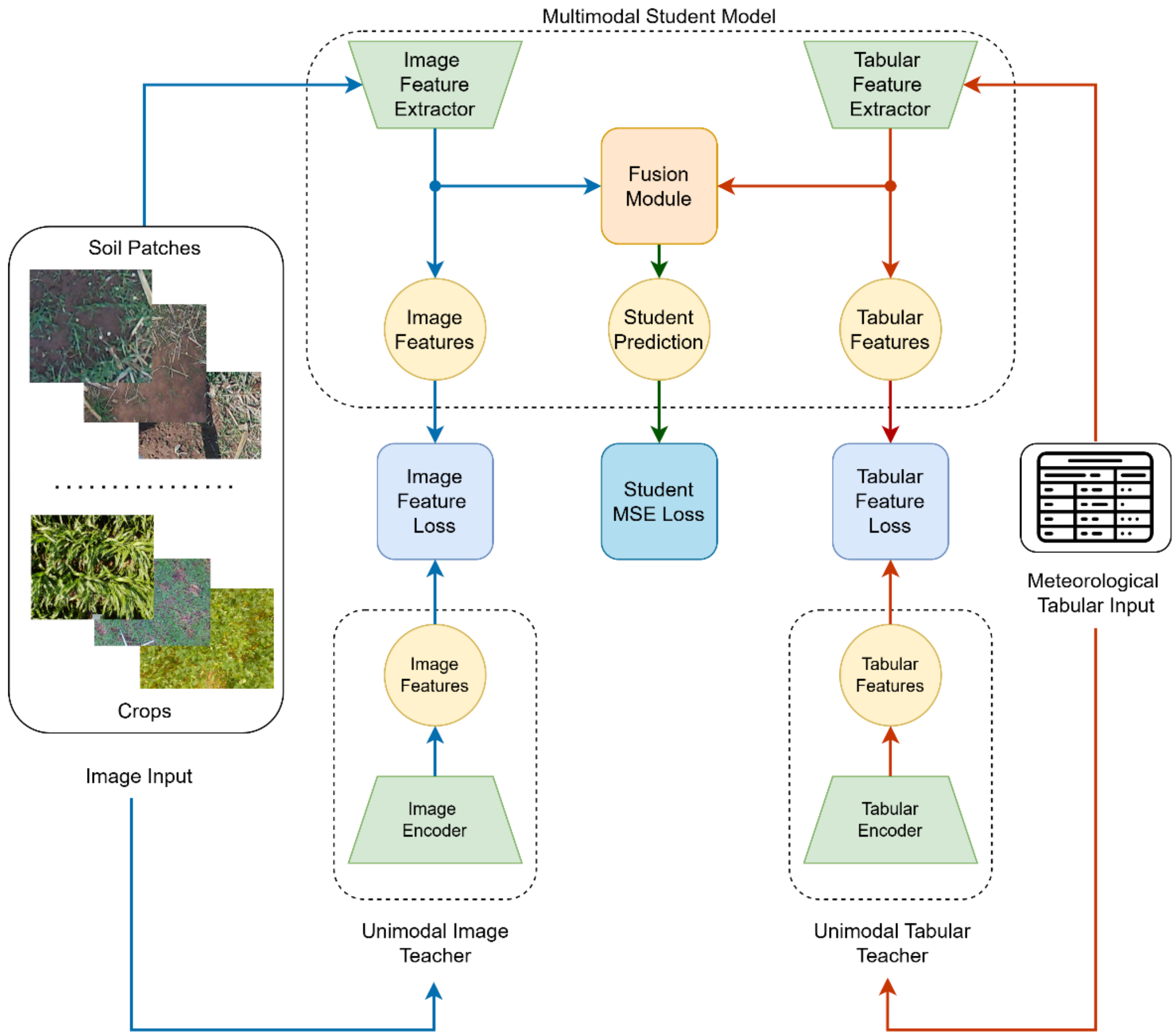

**Figure 1: Overview of our Methodology**. Soil patches or crop images, along with meteorological tabular data, serve as inputs to both the pretrained unimodal teacher models and a multimodal student model. The student aligns its unimodal encoders with the teacher models through feature-based losses while also optimizing for prediction accuracy using MSE loss against ground truth labels.

To address this challenge in soil moisture estimation, we propose Soil Moisture Distillation (SMD), a multimodal Knowledge Distillation (KD) framework that ensures both meteorological and visual data contribute effectively to the prediction task. Originally introduced for model compression (Bucila, Caruana and Niculescu-Mizil, 2006; Hinton, Vinyals and Dean, 2015), KD has evolved into a broader technique for transferring knowledge from a stronger teacher model to a student model, improving its learning efficiency. Multimodal KD extends this concept by leveraging information from multiple modalities to enhance representation learning, which supports several real-world AI applications (Wang *et al.*, 2023; Yuan and others, 2024). By leveraging unimodal teacher models, trained separately on meteorological and image data, we guide a multimodal student model to align its feature representations with those of the teachers. This structured supervision ensures that both modalities contribute meaningfully, mitigating modality imbalance and leading to improved soil moisture prediction.

To evaluate SMD, we conduct extensive experiments on the MIS-ME (Rakib *et al.*, 2024) dataset, comparing our KD-based approach against standard multimodal fusion techniques and state-of-the-art baseline models. Our results demonstrate that SMD outperforms traditional fusion methods and baselines. Additionally, we analyze the relative contributions of soil patches and crop images, showing that crop images provide richer visual cues, leading to superior predictive performance.

Our key contributions are as follows:

- We propose Soil Moisture Distillation (SMD), a KD-based multimodal framework that leverages unimodal teachers to enhance soil moisture prediction.
- We analyze the impact of soil patches versus crop images, demonstrating that crop images offer superior predictive performance.
- SMD achieves a 2% improvement in MAPE for soil patches and 1.2% for crops, surpassing baseline multimodal methods.

## METHODOLOGY

### Motivation

Integrating data from multiple modalities has been shown to significantly enhance model performance by capturing complementary information that is otherwise unavailable in a single modality. In the context of soil moisture estimation, combining meteorological tabular data and image-based features (e.g., crop images or soil patches) allows models to leverage both numerical patterns and visual cues, leading to more accurate predictions. This fusion of heterogeneous data enables the model to better generalize across diverse environmental conditions and reduces reliance on any single source of information.

However, multimodal learning often suffers from modality imbalance, where certain dominant modalities contribute disproportionately to the learning process while weaker modalities are underutilized. This imbalance can degrade model performance, as the network may prioritize features from a single modality instead of effectively integrating all available information. Multimodal Knowledge Distillation (KD) addresses this issue by employing unimodal teacher models, which are pretrained on individual modalities. These teachers guide the multimodal student model to align its features with unimodal expert representations, ensuring that each modality is utilized effectively. This structured supervision mitigates modality imbalance and enhances robustness.

For soil moisture estimation, the combination of meteorological tabular data (e.g., temperature, humidity, precipitation) and image data (e.g., soil/crop textures) is crucial for accurate forecasting. By distilling knowledge from unimodal teachers, SMD ensures that both meteorological and visual information contribute meaningfully to the prediction task, ultimately improving estimation accuracy and model interpretability.

### Problem Formulation

We consider a supervised learning setup where the goal is to estimate soil moisture content given multimodal inputs consisting of meteorological tabular data and image data. Formally, let $m$ denote the number of available modalities, where each modality has an associated unimodal teacher model $T_i$ that is trained independently. The multimodal student model $S$ is designed to integrate these modalities while benefiting from teacher guidance.

Each input sample consists of: tabular meteorological data ($X_{tab}$) and image data ($X_{img}$), which can be either soil patches or crop images. The unimodal teacher models process their respective inputs to produce predictions:

$$T_{\text{tab}}(X_{\text{tab}}) = Y_{\text{tab}}, \quad T_{\text{img}}(X_{\text{img}}) = Y_{\text{img}} \tag{1}$$

Similarly, the multimodal student model extracts modality-specific features $F_{\text{tab}}$ and $F_{\text{img}}$ before applying a fusion strategy:

$$F_S = f\left(F_{\text{tab}}, F_{\text{img}}\right) \tag{2}$$

where $f$ represents the chosen fusion technique, which can be any fusion method concatenation, late fusion, summation, etc.). In our case, we use concatenation. The final prediction is then computed as:

$$Y_S = g(F_S) \tag{3}$$

where $g$ is the output prediction function.

Figure 1 illustrates this pipeline, where both unimodal teacher models extract features independently from their respective inputs. The multimodal student model then integrates these features through the fusion module, producing a final multimodal output. Additionally, to compare the student model with the unimodal teachers, the unimodal outputs of the student model are derived by passing pre-fusion modality-specific features through dedicated linear layers. This setup allows direct supervision from each teacher model, ensuring that knowledge from each modality is preserved.

#### *Multimodal Student Loss*

Since soil moisture estimation is a regression problem, the multimodal student is trained using Mean Squared Error (MSE) loss between its predictions and the ground truth labels:

$$L_{\text{MSE}} = \frac{1}{N}\sum_{i=1}^{N}\left(Y_S^{(i)} - Y_{\text{GT}}^{(i)}\right)^2 \tag{4}$$

Where $Y_S$ is the multimodal student's prediction and $Y_{GT}$ is the true soil moisture value. The rationale behind this loss function is to minimize the squared differences, penalizing large deviations and ensuring accurate regression modeling.

*Feature Distillation Loss*

To enforce alignment between the multimodal student and unimodal teachers, we introduce a feature distillation loss for each modality:

$$L_{\mathrm{FD}} = \sum_{i=1}^{m} |F_S^{(i)} - F_T^{(i)}|^2 \tag{5}$$

where $F_S^{(i)}$ and $F_T^{(i)}$ are the feature representations of the student and the corresponding teacher for modality $i$.

The final training objective is a weighted sum of the MSE loss and feature distillation loss:

$$L_{\mathrm{total}} = \alpha\, L_{\mathrm{MSE}} + \beta\, L_{FD} \tag{6}$$

where $\alpha$ and $\beta$ are hyperparameters balancing the contribution of the two losses. For simplicity, in our experiments, we set $\alpha$ = 1.0 and $\beta$ = 1.0, treating both losses as equally important.

By aligning student features with their unimodal teacher counterparts, this approach ensures that each modality contributes effectively to the final prediction, mitigating the modality imbalance problem and leading to superior generalization in soil moisture estimation.

## EXPERIMENTS & RESULTS

### Experimental Setup

We evaluate SMD using the dataset introduced in the MIS-ME paper, which includes meteorological tabular data along with two types of image data: soil patches and crop images. Following the same 65:15:20 split for training, validation, and testing, we predict volumetric water content (VWC), measured in $cm^3cm^{-3}$. To investigate the impact of different image modalities, we integrate meteorological data once with soil patches and once with crops, comparing their performance to determine which modality contributes more effectively to soil moisture estimation.

We compare SMD with two baseline approaches: (1) Normal Fusion, which employs standard concatenation of features, and (2) the Hybrid Loss method proposed in the MIS-ME paper, which achieved the best results in their study. For consistency, we adopt the hyperparameter settings from the MIS-ME paper. For the image feature extractor, we use ResNet18 as the backbone for both the teacher and the student models. Similarly, for the meteorological feature extractor, we employ the fully-connected neural network architecture described in the MIS-ME paper for both teacher and student networks. This ensures a fair comparison between our KD approach and the baselines. We evaluate our method using Mean Absolute Error (MAE), Mean Absolute Percentage Error (MAPE), and $R^2$ score, ensuring a comprehensive assessment of model performance.

### Comparing Our KD Method with Baselines

Table 1 presents a comparison of SMD against baseline approaches for both soil patches and crops. Our method outperforms all baselines, demonstrating its effectiveness in soil moisture estimation. For soil patches, SMD improves the MAPE score by approximately 2% compared to both the normal fusion and MIS-ME methods, which exhibit similar performance. For crops, SMD achieves a 1.2% improvement in MAPE over MIS-ME and around 2.5% over normal fusion. These results highlight the efficacy of our KD framework in integrating multimodal data.

Additionally, when examining unimodal performances, we observe that the integration via our KD method yields a significant improvement in soil moisture estimation. Compared to unimodal models, the combination of modalities using our KD approach results in better generalization and more accurate predictions. This demonstrates that unimodal teachers effectively guide the multimodal student, leading to more robust feature representations.

Finally, we note that crop images consistently outperform soil patches across all methods. This may be attributed to the fact that crop images capture a broader horizon, encompassing both soil properties and crop characteristics, thereby providing a more comprehensive context for soil moisture estimation. In contrast, soil patch images focus solely on the soil surface, offering limited information that may not fully represent the dynamic interactions between soil and vegetation.

**Table 1: Result Comparison**

| Baselines | Soil Patches | | | Crops | | |
|---|---|---|---|---|---|---|
| | MAE | MAPE | R2 | MAE | MAPE | R2 |
| Meteorological Model | 0.035 | 13.080 | 0.485 | 0.040 | 15.495 | 0.321 |
| Image Model | 0.037 | 14.227 | 0.360 | 0.017 | 6.742 | 0.819 |
| Normal Fusion | 0.039 | 15.044 | 0.383 | 0.019 | 7.188 | 0.811 |
| MIS-ME | 0.039 | 15.163 | 0.395 | 0.015 | 5.928 | 0.865 |
| SMD (Ours) | **0.034** | **13.022** | **0.491** | **0.012** | **4.751** | **0.901** |

### Residual Analysis

Residual analysis evaluates prediction accuracy by examining the deviation between predicted and actual soil moisture values. Figure 2 compares residual distributions for normal fusion and SMD using soil patches and crops. A higher percentage of residuals within ±0.05 $cm^3/cm^3$ indicates better alignment between predictions and ground truth.

For soil patches, normal fusion captures 67.04% of residuals within ±0.05 $cm^3/cm^3$, while SMD improves this to 75.88%, demonstrating reduced prediction error and improved generalization. The wider spread in normal fusion residuals suggests higher uncertainty, whereas SMD produces a more centered and consistent distribution.

For crops, normal fusion achieves 93.11%, whereas SMD further improves accuracy to 95.89%. This confirms that crops provide better predictive performance than soil patches and that SMD enhances feature extraction by leveraging unimodal teachers, reducing residual variance.

Overall, our framework significantly improves prediction accuracy, particularly for soil patches where normal fusion struggles with higher residual errors. By aligning student features with unimodal teachers, SMD refines feature representations and reduces uncertainty, making it a more robust framework for soil moisture estimation.

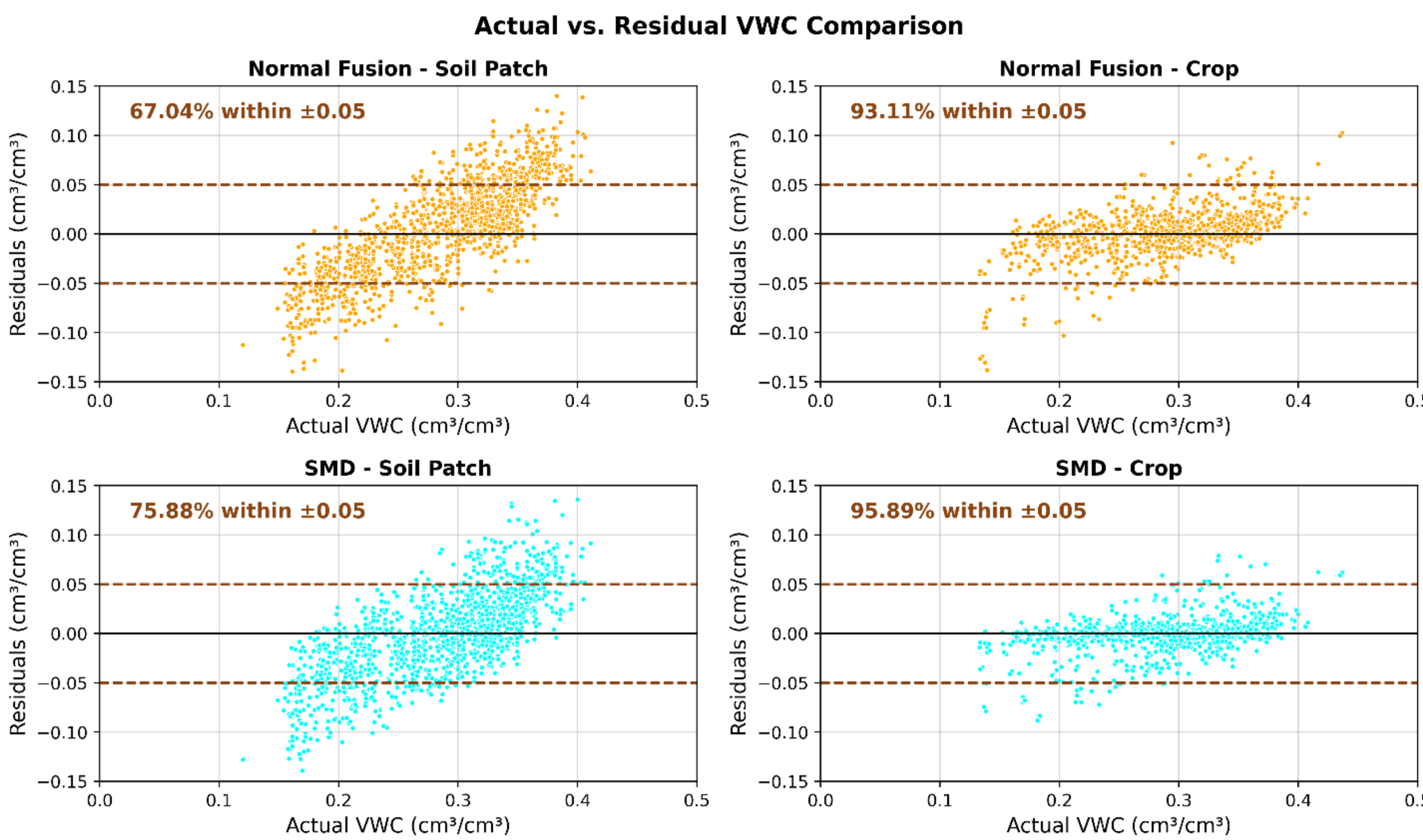

**Figure 2: Residual Comparison between Normal Fusion and SMD**

## CONCLUSION

In this work, we proposed Soil Moisture Distillation (SMD), a Knowledge Distillation (KD)-based multimodal learning framework for soil moisture estimation by integrating image and meteorological data. SMD addresses the modality imbalance problem by leveraging unimodal teacher models, ensuring effective utilization of both modalities. By aligning the multimodal student model with unimodal expert representations, we enhance feature extraction and improve soil moisture regression. Additionally, our analysis of soil patches and crop images highlights the importance of incorporating visual cues alongside meteorological data for a more comprehensive estimation framework.

For future work, we aim to explore KD for model compression, making the SMD models more lightweight and efficient for deployment on edge devices. Optimizing the model for low-power agricultural systems will enable real-time soil moisture prediction in resource-constrained environments, facilitating broader adoption of AI-driven precision agriculture solutions.

## ACKNOWLEDGEMENTS

This work is funded by the U.S. Department of Agriculture, National Institute of Food and Agriculture, Data Science for Food and Agricultural Systems (DSFAS) program. Accession number 1030694.